\documentclass[letterpaper, 10 pt, conference]{ieeeconf}  

\IEEEoverridecommandlockouts                              

\usepackage{graphicx}
\usepackage{amsmath}
\usepackage{multirow}
\usepackage{caption}
\usepackage{tabularx}
\usepackage{amssymb}
\usepackage{balance}

\begin{document}
	
\title{\LARGE \bf
	A Fine-Grained Visual Attention Approach for Fingerspelling Recognition in the Wild}


\author{{Kamala Gajurel$^{\dagger}$, Cuncong Zhong$^{\dagger}$, Guanghui Wang$^{\ddagger}$}\\
	{$^\dagger$ \textit{Department of Electrical Engineering and Computer Science, University of Kansas, Lawrence KS, USA}}\\
	$^\ddagger$ \textit{Department of Computer Science, Ryerson University, Toronto ON, Canada, M5B 2K3}\\
}

\maketitle
\thispagestyle{empty}
\pagestyle{empty}
	
\begin{abstract}

Fingerspelling in sign language has been the means of communicating technical terms and proper nouns when they do not have dedicated sign language gestures. Automatic recognition of fingerspelling can help resolve communication barriers when interacting with deaf people. The main challenges prevalent in fingerspelling recognition are the ambiguity in the gestures and strong articulation of the hands. The automatic recognition model should address high inter-class visual similarity and high intra-class variation in the gestures. Most of the existing research in fingerspelling recognition has focused on the dataset collected in a controlled environment. The recent collection of a large-scale annotated fingerspelling dataset in the wild, from social media and online platforms, captures the challenges in a real-world scenario. In this work, we propose a fine-grained visual attention mechanism using the Transformer model for the sequence-to-sequence prediction task in the wild dataset. The fine-grained attention is achieved by utilizing the change in motion of the video frames (optical flow) in sequential \textit{context-based attention} along with a Transformer encoder model. The unsegmented continuous video dataset is jointly trained by balancing the Connectionist Temporal Classification (CTC) loss and the maximum-entropy loss. The proposed approach can capture better fine-grained attention in a single iteration. Experiment evaluations show that it outperforms the state-of-the-art approaches.     
\end{abstract}

\begin{keywords}
Fingerspelling recognition; visual attention; Transformer network.
\end{keywords}

\section{INTRODUCTION}


Sign language is the most structured form of non-verbal communication for people with hearing and speaking deficiency. It incorporates the symbols or gestures made by hand including facial expressions and postures of the body. There are over 300 different sign languages and about 70 million deaf people around the world according to the report of the World Federation of the Deaf \cite{statAsl}.
The sign language dialects vary from place to place although some of the gestures may overlap.
One of the families of sign language is the American Sign Language (ASL). It is the natural language for around 500,000 people in the US and Canada \cite{ASL}. ASL is also the third most widely studied language in the United States in addition to English according to the Modern Language Association's study of US colleges and Universities \cite{looney-18enrollments}. ASL has its own distinct syntax and grammar which change and develop over time. 

\begin{figure}[t]
\begin{center}
    \includegraphics[width=250mm, height=5cm,keepaspectratio]{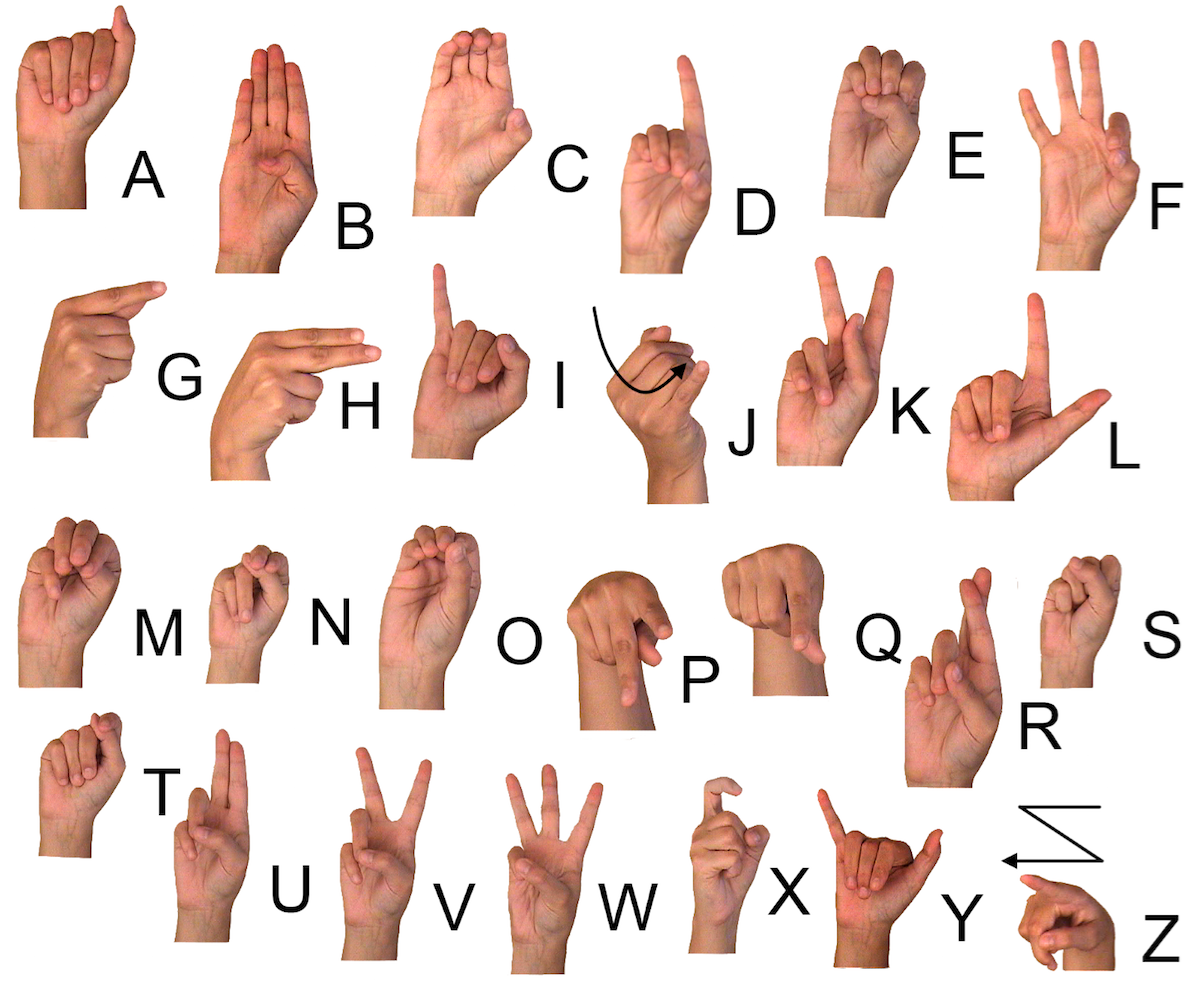}
\end{center}
\caption{The symbols of alphabets in ASL Fingerspelling \cite{Fingerspelling}}\label{fig:asl}
\end{figure}

One of the most important elements of sign language is fingerspelling. Fingerspelling is practical but constrained component of sign language \cite{rastgoo-2020sign}. It is the representation of the individual letters and numbers using standard finger positions. Fingerspelling is generally used when there are no distinct signs for the words \cite{shi-19fingerspelling}. These words are either technical words or proper nouns, such as personal names, places' names, brands, etc.  Although there exist signs for words, the fingerspelling can be employed to emphasize them in focus construction \cite{montemurro-18emphatic}. Fingerspelling covers 35\% in ASL and is often used in social interaction or conversations involving technical events \cite{padden-03alphabet}. Fig. \ref{fig:asl} shows the symbols of alphabets used in ASL.

Sign language recognition is the task of classifying the symbols or gestures made by hand along with the facial expressions and postures of the body. Two individuals can easily understand the symbols or gestures made by each other if both have a well understanding of the language. However, the same symbol cannot be made comprehensible if one of the communicators is a person with limited knowledge of the language or a machine. Sign language recognition has been a very important means to reduce the communication barrier for deaf individuals \cite{shi-19fingerspelling}. Furthermore, there are many benefits of sign language recognition in applications such as assistive technologies for the deaf, interpreting services, translation services, human-computer interaction, as well as robotics \cite{rastgoo-2020sign}.

Fingerspelling recognition, a subtask of sign language recognition, is essential for communication in specialized areas and online social media for the deaf \cite{shi-19fingerspelling}. Fingerspelling recognition can also assist in the teaching and learning process for beginners. Fingerspelling in ASL is relatively simple in comparison to the general sign language recognition task since fingerspelling includes only one hand to make symbols most of the time \cite{shi-19fingerspelling}. This work is mainly focused on the fingerspelling recognition task.

Although fingerspelling recognition only involves one hand with limited vocabulary, it remains a very challenging task. The main challenge in fingerspelling recognition is the large inter-class similarity and large intra-class variation in alphabets leading to incorrect classification. The difficulty in recognition also depends upon the setting where the dataset is prepared. The recent increase in deaf online media has motivated recognition in the wild dataset which reflects the real-world scenario. In such data, the difficulty arises due to the motion blur of low-quality video \cite{shi-19fingerspelling}. Other challenges that exist in the recognition task are the variability in the appearance of signers' bodies and the large number of degrees of freedom \cite{rastgoo-2020sign}, diverse viewpoint and orientation of the hand \cite{zimmermann-17learning}, as well as the difficulty in capturing the spatial and temporal context during fingerspelling.

It is essential to obtain the fine-grained features of the hand to handle the challenges associated with fingerspelling recognition. The signing hand should be localized to identify the region of interest by focusing on the specific spatial region. Further refinement in the region of interest should be done to capture the temporal context as well as the subtle differences between the gestures.

State-of-the-art for capturing fine-grained features in fingerspelling recognition is based on the spatial attention with recurrent neural network \cite{shi-19fingerspelling}. This paper implements the iterative visual attention that helps in obtaining the regions of interest with high resolution. Our proposed model introduces the context-based attention mechanism to refine the attention weights, followed by the Transformer encoder model. This model also implements the maximum entropy loss in the field of fingerspelling recognition for regularization. Our model outperforms the state-of-the-art performance in a single iteration by around 3\%. Moreover, the computation time for training and inference are significantly reduced due to the use of the Transformer model and single iteration. The source code of the proposed model will be available on the author's homepage \footnote{https://github.com/rucv/Sequence-Fingerspelling}.

The rest of the paper is organized as follows. In the next section, we describe the related work in the domain of fingerspelling recognition. Section \ref{probform} formally introduces the sequence-to-sequence prediction problem in fingerspelling recognition. Section \ref{method} describes the proposed model. The experiments conducted and the results are described in sections \ref{exp}, and \ref{result}, respectively. Finally, we conclude the paper in section \ref{conc}. 

\section{Related Work}
With the fast development of deep learning, convolutional neural networks have been successfully applied to various classification and recognition tasks \cite{cen2019boosting, li2021sgnet, patel2020comparative, sajid2021parallel, xu2020adaptively, zhang2020efficient}. With the availability of the large-scale image/video dataset in online media such as youtube and sign language websites, many sign language recognition approaches using the RGB-based data have been proposed. In this work, we focus on the sequence-to-sequence prediction task with the fingerspelling videos. A number of prior works have been conducted to address this problem, taking videos as input and letter sequence as output \cite{kim2017lexicon, shi2017multitask, shi2018american, liwicki2009automatic, papadimitriou2019end, shi-19fingerspelling}.


Our work is closely related to two studies \cite{shi2018american} and \cite{shi-19fingerspelling}. The paper \cite{shi2018american} introduces the naturally occurring video dataset \textit{ChicafoFSWild} collected from \textit{YouTube, aslized.org, and deafvideo.tv}. This paper first segments out the signing hand using the hand detector model from the image frames. The images so obtained are then passed through the CNN layer for feature extraction. Then the features are fed through long short-term memory (LSTM). By using either connectionist temporal classification (CTC) or attention-based LSTM decoder, the prediction of the sequence is performed. The result shows that CTC based prediction model gives better performance in comparison to decoder-based. 

Paper \cite{shi-19fingerspelling} works on the same dataset and it has introduced another larger dataset \textit{ChicagoFSWild+} for further experimentation. The main idea of this paper is to use an end-to-end model based on an attention-based mechanism. This designed model does not explicitly localize the hand as done in \cite{shi2018american}. By using the iterative attention mechanism followed by the CTC model, this paper improves the performance of \textit{ChicagoFSWild}. In addition, the performance in test data based on training using both wild and wild+ data shows that the result in test data improves by a significant amount. The proposed work uses Transformer based attention mechanism instead of an RNN-based spatial attention mechanism.

Another work done in \textit{ChicagoFSWild} is by the paper \cite{parelli2020exploiting}. This paper uses 3D hand pose estimation. The evaluation shows that the performance in the test data increases slightly in comparison to \cite{shi-19fingerspelling}.
The paper \cite{camgoz2020sign} has proposed the Transformer-based encoder-decoder architecture for sign language recognition. Our proposed model differs from the previous study and we focus on fingerspelling recognition using context-based attention followed by the Transformer encoder only.

\section{Problem Formulation}\label{probform}
Sequence-to-sequence prediction with videos falls under a challenging task in fingerspelling recognition since it involves the video dataset with continuous multiple signs in a single video. Given a set of images representing video frames  $I_1$, $I_2$...,$I_T$, the task is to predict the target letter sequence $l_1$, $l_2$,....,$l_K$ where $K \le T$. The labeling in this task is unsegmented labeling with no alignments between the frames and labels. Connectionist Temporal Classification (CTC) \cite{graves2006connectionist} is used to find the predicted letter sequence in such a setting. 

For given video frames $I^{1:T}$, the first step for the CTC model is to find a function $f$ that learns the probability distribution of labels for each frame $y^{1:T}$.
\begin{equation}
y^{1:T} = \sigma(f(I^{1:T}, \theta))
\end{equation}
where $\sigma$ is a softmax function, $\theta$ is the model parameters, and $y \in [0,1]^{C'}$ contains the probabilities of $C$ classes  and one extra class blank which indicates none of the classes. These output probabilities define the different possible alignments of label sequences with the input sequence. The probability of an alignment $\pi$ for a given sequence is defined as follows:
\begin{equation}p(\pi|\mathbf{x}^{1:T}) = \prod_{t=1}^T{y_{\pi_t}^t}
\end{equation}
where $\pi \in C'^T$ and $\mathbf{x}$ is the visual features obtained from $f$.

Next, a many-to-one map $\mathbf{B}:C'^T \mapsto {C^{\le T}}$ is defined by removing all repeated labels and blanks from the alignment. For example: $\mathbf{B}(aa-ss-l) = \mathbf{B}(-a-sll) = asl$. The path with same character in adjacent position is handled by a blank in between the repeat characters. For example: $\mathbf{B}(ee-ggg-g) = egg$. The map $\mathbf{B}$ is used to define the conditional probability of letter sequence $\mathbf{l} = l_1, l_2..,l_k$ given visual features $\mathbf{x}$.
\begin{equation}
p(\mathbf{l}|\mathbf{x}) = \sum_{\pi \in \mathbf{B}^{-1}(\mathbf l)} p(\pi|\mathbf{x})
\end{equation}

The CTC model is used to optimize this probability for the ground-truth label sequences by using CTC-loss.
In addition to the CTC-loss, maximum entropy loss (MEL) \cite{dubey2018maximum} is also applied as a regularization to avoid overfitting.
\begin{equation}\label{eq:MEL}
\text{MEL} = \log_2(C')- {\frac{1}{T} \sum_{t=1}^T \sum_{k=1}^{C'}-p(y^t_k)\log_2(p(y^t_k))}
\end{equation}
where $T$ is total video frames, $y_k^t$ is the probability of class $k$ for frame $t$. The goal of using MEL is to encourage the probability distribution that does not spike toward single class.

\section{Methodology}\label{method}
\begin{figure}[!t]
    \centering
    \includegraphics[width=\linewidth, keepaspectratio]{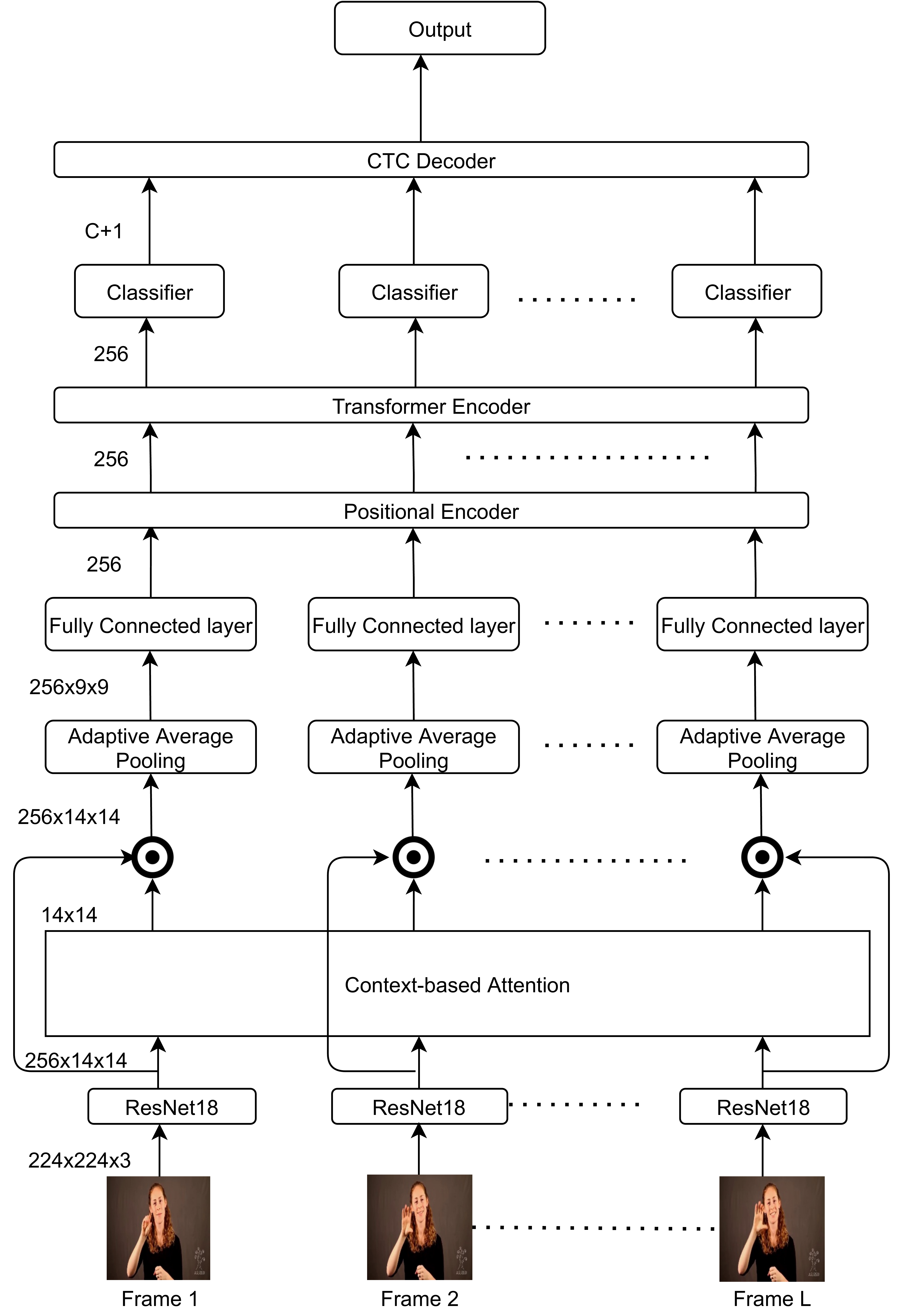}
\caption{Block diagram for the sequence-to-sequence prediction of video dataset}\label{fig:ModelVideo}
    \end{figure}

The proposed approach for sequence-to-sequence prediction is depicted in Fig. \ref{fig:ModelVideo}. The video dataset is first converted to the sequence of frames and the frames are processed to obtain the dimension of 224x224x3. The frames are passed through the pre-trained ResNet18 parallelly to extract the feature maps with dimensions 256x14x14 for each frame. The output of ResNet18 is passed through the context-based attention module, described in section \ref{contextbased}, to obtain the spatial attention weights for each frame. 

The context-based attention weights are multiplied with the feature maps obtained from ResNet18 to produce the attended feature maps.
The attended feature maps of each frame are then passed through adaptive average pooling. This pooling method converts each frame to the size of 256x9x9. The output from the pooling layer is now used by the fully connected layer to summarize the features into 256-dimensions image embedding for each frame. The summarized embedding along with the positional information obtained from the positional encoder \cite{vaswani2017attention} is now passed through the transformer encoder layer. The Transformer encoder layer generates the context and relevant information from the input. After the encoder layer, the 256-dimensional encoding is transferred to the classifier. The classifier is a fully connected layer that produces the output of C+1 dimensions. Here, C stands for the number of classes and one extra class is for \textit{blank} character that represents the transition of signs. The log probabilities of each class are now passed to the Connectionist Temporal Classification (CTC) decoder to predict the letter sequence of the given video as the output. 
    
     \begin{figure}[!t]
    \centering
    \includegraphics[width=\linewidth, keepaspectratio]{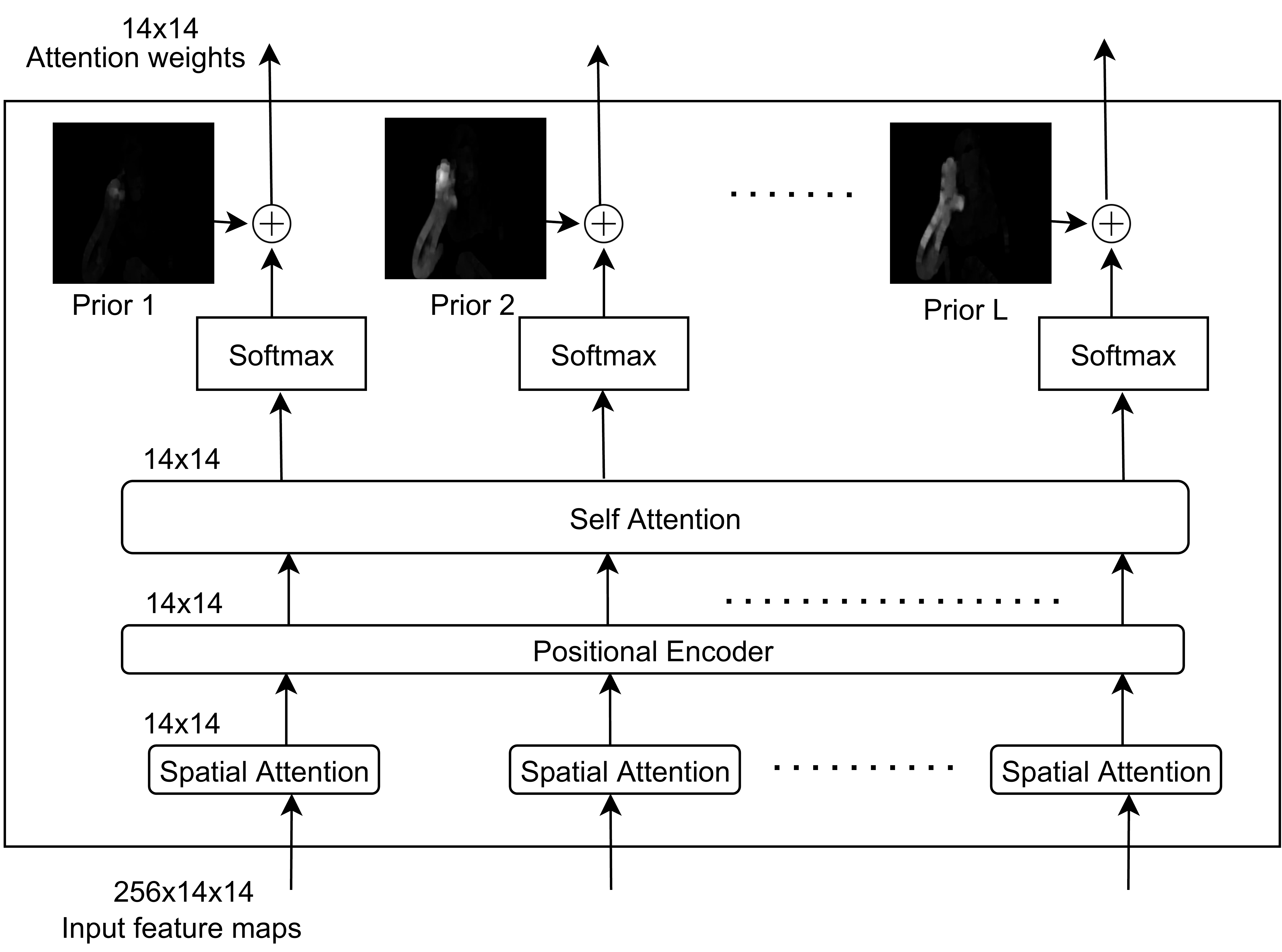}
\caption{Block diagram for the context-based attention module.}\label{fig:contextAtt}
    \end{figure}

\subsection{Context-based Attention Module}\label{contextbased}
Context-based attention shown in Fig. \ref{fig:contextAtt} comprises of spatial attention layer, positional encoder and self-attention layer along with the priors ($P$). First, the feature map of each frame is processed to obtain the corresponding spatial attention map ($\mathbf{A_i}$).

\begin{equation}
    \mathbf{A_i} = \delta[\delta(\mathbf{F_i}\times W_a)\times W_v], \forall i \in \{1,..,T\}
\end{equation}
where, $\mathbf{F_i} \in \mathbb{R}^{14\times14\times256}$ is the transposed feature map, $W_a \in \mathbb{R}^{256\times512}$ and $W_v \in \mathbb{R}^{512\times1}$ are the attention weights learned using back-propagation, and $\delta$ is the ReLU activation function. 

The spatial attention layer does not take the context of the prior frames. So, a positional encoder followed by a self-attention layer is used. The positional encoder is used to get the relative positional information from all the frames. The self-attention layer \cite{vaswani2017attention} is then applied to this attention map ($\mathbf{A} \in \mathbb{R}^{L\times 14\times 14}$) from the spatial attention layer to obtain the refined map $\mathbf{A^s}$.  
 
Next, the weighted sum of the prior and the attention weights of each frame obtained from the self-attention layer followed by softmax is calculated to find the final attention weight of $14 \times 14$ dimension. 
\begin{equation}
\mathbf{A^p_i} = w_{p}\times P_i + (1-w_p)\times \sigma(\mathbf{A^{s}_i})
\end{equation}
where $\mathbf{A^p_i}$ is the final attention weight and $w_p$ is the weight for prior learned by the model. Prior is generated using optical flow \cite{farneback2003two} using the same process followed by \cite{shi-19fingerspelling}. 

The final attention is now multiplied with the feature map from ResNet18 to obtain the attended feature map.
\begin{equation}
\mathbf{F_i} = \mathbf{F_i} \odot \mathbf{A^p_i}
\end{equation}
where $\mathbf{F_i}$ is the feature map of each frame.

\subsection{Transformer Encoder}
The Transformer encoder \cite{vaswani2017attention} consists of a multi-head attention layer, feed-forward network along with add and normalize layer. The multi-head attention layer helps the model to jointly attend to the information from different positions of embeddings. The multi-head attention used here is masked multi-head attention so that each embedding can take the position information from previous frame embeddings only for refining the attention in each frame. The attention layer is followed by the add and normalize layer which acts as the residual connection between two sub-layers. The feed-forward network in the transformer encoder processes the attended embedding from the multi-head attention layer independently and identically.

\subsection{CTC Decoder}
We consider the following three types of CTC decoding techniques.
\begin{itemize}
    \item \textbf{Greedy decoding.}
Greedy decoding is the method in which the decoder chooses the letter with maximum probability for each frame. Mathematically,
\begin{equation}
\hat{l}_t = argmax_k (y^t_k) , \forall k \in \{1,....,C+1\}
\end{equation}
where $\hat{l}_t$ is the predicted letter and $y^t_k$ is the probability distribution of each class k.

Although the greedy decoding method is fast, it chooses the alignment path with the highest probability as the solution, while there exist many alignment paths that merge into a single sequence label. For example: let 'cat' be the real sequence with 5 frames. There may be alignments such as \texttt{-oat-}, \texttt{-ccat}, \texttt{--cat}, \texttt{cc-at}, and \texttt{c-aat} with probabilities $0.4$, $0.2$, $0.2$, $0.1$, and $0.1$, respectively. In this case, if the greedy method is employed, the prediction will be \texttt{oat}. However, other alignments collapse to \texttt{cat} with a probability of $0.6$. Thus, it is better to use a beam search method that mitigates this problem.
\item \textbf{Beam search decoding.}
Beam search decoding in CTC expands all the possible next states and retains `BW' (beam width) most likely states as the sequence is constructed. The retaining is done after collapsing the alignments not ending with blank. Moreover, the probability scores are merged for those alignments. BW is the hyper-parameter that balances the decoding speed and accuracy of performance. 

\item \textbf{Beam Search with Language Model.}
The Character-level Language model aims to predict the next character given the previous characters in the sequence. Integrating language model in CTC decoder can boost the performance of the model by capturing the relationship between the characters for the given domain. For the first time step, the language model uses the observed marginal probabilities of the characters. For the subsequent characters, the language model considers the conditional probabilities of the character given previous $k$ characters. The value $k$ is also known as the order of the language model.

The language model can be integrated with the beam search decoding by modifying the scores of the sequences obtained after merging the same states. Let $s_b$ be the score of beam search and $\alpha$ be the weight of the language model. Then the score after applying the language model is 
\begin{equation}
s = (1-\alpha)\times s_b + \alpha \times P(\hat l_t|\hat l_{t-1},...,\hat l_{t-k})
\end{equation}
where $\hat l_t$ is the predicted character at position $t$.
\end{itemize}

\section{Experiments}\label{exp}
\subsection{Dataset}
The proposed model is tested on the ChicagoFSWild \cite{shi2018american} dataset. The dataset consists of the fingerspelling clips from ASL videos available on online media such as Youtube, aslized.org, and deafvideo.tv. There are 7,304 clip sequences in total. The data is split into 5,455 training, 981 validation (development), and 868 test sequences. The sequences are mostly right-handed (6,782 sequences), a few left-handed (522 sequences), and rarely both-handed (121 sequences). The dataset consists of 192 unique signers among which 91 are male and 77 are female. Each unique signer is assigned to only one of the data partitions for the signer independent setting. The vocabulary consists of 26 fingerspelling alphabets (a-z) and 5 other special characters (SPACE, \&, ', ., @). The special characters occur very rarely.

\subsection{Data Preprocessing}
\begin{itemize}
        \item \textbf{Whole frame} The original raw video frames in the wild dataset are of different dimensions since the videos are collected from different online sites. To convert the frames into a uniform dimension, the frames are resized to the dimension of 224 $\times$ 224. The resized frames are normalized using the mean [0.485, 0.456, 0.406] and std = [0.229, 0.224, 0.225]. The priors obtained from optical flow are resized to the feature map dimension of 14 $\times$14. 

\item \textbf{Face ROI} In general, the fingerspelling is carried out such that the signing hand is near the face region. This motivates using the existing face detection approaches for obtaining the region of interest (ROI). The data preprocessing for Face ROI is followed from the paper \cite{shi-19fingerspelling}. After obtaining the frames with the region of interest, they are also resized to the dimensions of 224 $\times$ 224 and normalized using the same mean and standard deviation used in the Whole frame case. Moreover, the priors are resized to 14 $\times$ 14.
\end{itemize}

\subsection{Models and Setup}
For our proposed model, the Transformer encoder uses 2 layers with masked 2-head attention. The masking is done in such a way that only 5 prior frames are considered to refine the attention in each frame. The number of hidden units for the feed-forward network is 1024. Dropouts of 0.3 and 0.1 are used for the Transformer layers and context-based attention layer respectively.
The following three ablation studies have been conducted.
\begin{itemize}
	\item \textbf{CTC loss:} In this approach, we use only CTC loss while training the model. The purpose of this approach is to observe how well the proposed model performs without regularization and augmentation. 
	
	\item \textbf{Maximum entropy loss:} This approach uses CTC loss along with maximum entropy loss as regularization for model training. CTC loss generally gives the impulsive or spike distribution leading to overfitting 
	\cite{liu2018connectionist}. To mitigate this problem, maximum entropy loss, described in equation \ref{eq:MEL}, is used as regularization. 
	
	\item \textbf{Horizontal flipping:} The input dataset distribution consists of both right and left signing hands. However, the distribution is imbalanced with left-handed signers being about 7 \% of the total samples. To generalize the model, some of the input training samples are flipped at random. This process of data augmentation alleviates the problem of imbalanced distribution of signing hands. The probability of flipping the input is taken as 0.3 after hyperparameter tuning. 
\end{itemize}
The experiments are performed for different combinations of CTC loss, maximum-entropy loss, and horizontal flipping.
All the experiments are implemented using the PyTorch framework \cite{NEURIPS2019_9015} using one NVIDIA Tesla P100 GPU. PyTorch's implementation of CTC-loss is used with AdamW optimizer \cite{loshchilov2017decoupled}. The learning rate of $10^{-4}$ is used by keeping other parameters of AdamW default. 

The experiments are implemented for Whole frame and face ROI preprocessing and are trained for 20 epochs.  
For each experiment, the letter accuracy is reported for greedy decoding, beam search decoding\footnote{https://github.com/parlance/ctcdecode} with a beamwidth of 20, and beam search with KenLM \cite{heafield-etal-2013-scalable} character-level language model decoding.  The language model weight ($\alpha$) is tuned to 0.2. The letter accuracy for a predicted sequence is calculated using the approach followed by \cite{shi2018american, shi-19fingerspelling}. The approach is to find the minimum edit distance between the predicted sequence and the ground truth letter sequence. Mathematically, 
\begin{equation}
 \textit{Letter accuracy} (L_{acc}) = \max(0, 1 - \frac{S+D+I}{N} )
\end{equation}
where S, D, I represent the substitution, deletion, insertion respectively in the alignment. N is the number of total letters in the ground-truth sequence. The $max$ in the equation ensures positive accuracy when $N < S+D+I$.

\section{Result and Discussion}\label{result}
Table \ref{tab:letterAccComp} shows the comparison of letter accuracy between the state-of-the-art models and the proposed model. The comparison is done on both the development (dev) and test data sets. As seen in the table, the proposed model outperforms both the development and test accuracy in comparison to the other two approaches. In both the proposed model and \cite{shi-19fingerspelling}, Face ROI is used. The proposed model is trained on a single iteration using CTC loss, maximum entropy loss, and horizontal flipping. \cite{shi-19fingerspelling} is an iterative visual-based attention model trained on three iterations using CTC loss only. In both cases, beam search decoding with a language model (BS+LM) is employed while evaluating. Paper \cite{parelli2020exploiting} has exploited and used the 3D hand pose information and reported the performance only in the test dataset as 47.93\%.
\begin{table}[!t]
\centering
\sffamily
\caption{Comparison of letter accuracy for continuous gestures on ChicagoFSWild benchmark dataset trained on ChicagoFSWild/train partition.}
\begin{tabular}{|l|c|c|}
	\hline
	\multirow{2}{5em}{\textbf{Experiment}} & \multicolumn{2}{c|}{\textbf{Letter Accuracy (\%)}}\\
	\cline{2-3}
	& \textbf{dev} & \textbf{test}\\
	\hline
	Shi et al. {\cite{shi-19fingerspelling}} & 46.8 & 45.1\\
    \hline
    Parelli et al. {\cite{parelli2020exploiting}} & -& 47.93\\
    \hline
     {Proposed Model} & \textbf{46.96}& \textbf{48.36}\\
    \hline
\end{tabular}
\label{tab:letterAccComp}
\end{table}

\begin{table}[!t]
\centering
\sffamily
\caption{Ablation study in development dataset of ChicagoFSWild based on different combinations of CTC loss, maximum entropy loss and horizontal flipping for Face ROI.}
\begin{tabular}{|p{10em}|c|c|c|}
	\hline
	\multirow{2}{5em}{\textbf{\text{Study Cases}}} & \multicolumn{3}{c|}{\textbf{Letter Accuracy (\%) in dev}}\\
	\cline{2-4}
	& \textbf{Greedy} & \textbf{BS}& \textbf{BS+LM}\\
	\hline
	CTC loss only & 38.20 & 38.76 & 39.50\\
    \hline
    CTC loss and Maximum Entropy loss & 39.79 & 42.02 & 42.20\\
    \hline
    CTC loss and Horizontal flipping & 42.52 & 43.56 & 44.44\\
    \hline
    CTC loss, Maximum Entropy loss and Horizontal flipping & 44.82 & 45.77 & 46.96\\
    \hline
\end{tabular}

\label{tab:Abstudy}
\end{table}

Similar to the results of \cite{shi-19fingerspelling}, it is observed that the performance for Face ROI is superior to that of Whole frame (33.68 \% using beam search decoding). Thus, we only report the ablation study for Face ROI case. Table \ref{tab:Abstudy} shows the performance of the proposed model trained on different settings and evaluated on three decoding techniques: greedy, beam search (BS), and beam search with language model (BS+LM). The evaluation is done on the development dataset only to find out the best model for testing on the test dataset.  From the ablation study, it is observed that the inclusion of both maximum entropy loss and horizontal flipping makes the model better. The best model outperforms other settings for all three decoding techniques. While comparing three methods of decoding, beam search decoding with the language model gives better performance.

The proposed hypothesis is that the maximum entropy loss helps in regularization and horizontal flipping data augmentation helps in the generalization of the model across data distribution. The purpose of the ablation study is to test this hypothesis.  From the table, both these methods help in improving the model performance on the unseen data. Moreover, the combination of these two approaches boosts the performance even further, outperforming the state-of-the-art in development dataset.

In the context of time taken for training the proposed model, the average training time is about 6 ms per frame using one NVIDIA Tesla P100 GPU. The study in \cite{shi-19fingerspelling} has reported the average training time as 65 ms per frame on an NVIDIA Tesla K40c GPU. The difference in time taken is partially due to the parallel training of Transformers as compared to the sequential LSTM.

\begin{table}[!t]
\centering
\sffamily
\caption{Comparison of letter accuracy for development dataset based on iterations}

\begin{tabular}{|c|c|c|c|}
\hline
\textbf{Model}&\textbf{Experiment}& \textbf{First Iter.}& \textbf{Final Iter.}\\
\hline
    Shi et al. \cite{shi-19fingerspelling}& Whole frame & 23.0\%(w/o LM) & 43.6\% \\
\hline
     Proposed Model& Whole frame & 33.68\%(w/o LM) & 36.38\%\\ 
\hline
    Shi et al. \cite{shi-19fingerspelling}& Face ROI & 35.2\% & 46.8\% \\
\hline
     Proposed Model& Face ROI & 46.96\%& 45.01\%\\ 
\hline
\end{tabular}

\label{tab:iterAccComp}
\end{table}

Table \ref{tab:iterAccComp} shows the comparison of performance in development data for the first iteration and final iteration. \cite{shi-19fingerspelling} reports the letter accuracy of final iteration for both Whole frame and Face ROI using beam search decoding with the language model. However, for the first iteration in the Whole frame, only the performance with the beam search decoder is reported. So, the table also reports the performance of the proposed model in a similar manner to maintain consistency.
According to the table, it is obvious that the performance of the proposed model in the first iteration outperforms in both the Whole frame and Face ROI case with a significant amount. However, for the Whole frame in the final iteration, the performance does not improve in comparison to the state-of-the-art although there is an improvement from the first to the final iteration. For the Face ROI case, there is a small difference in the performance in comparison to the first iteration. 

 
\section{Conclusion}\label{conc}
In this paper, we have proposed a fine-grained visual attention approach for the sequence-to-sequence prediction task of fingerspelling recognition. We exploit the Transformer-based contextual attention mechanism for capturing the fine-grained details. The maximum entropy loss helped in the regularization and horizontal flipping data augmentation assisted in the generalization of the model. The proposed Transformer-based visual attention model significantly improved state-of-the-art performance in the ChicagoFSWild dataset. However, it is still low compared to that of human performance. We will further evaluate our model on a new large dataset ChicagoFSWild+, collected with crowdsourced annotators. The work could also be extended to end-to-end fashion for hand segmentation. 





\section*{ACKNOWLEDGMENT}
The work was supported in part by the Natural Sciences and Engineering Research Council of Canada (NSERC) under grant number RGPIN-2021-04244.


\balance
\bibliographystyle{IEEEtranS}
\bibliography{reference.bib}{}

\end{document}